\documentclass[10pt,twocolumn,letterpaper]{article}

\usepackage[pagenumbers]{cvpr} 

\usepackage{graphicx}
\usepackage{amsmath}
\usepackage{amssymb}
\usepackage{booktabs}
\usepackage{multirow}
\usepackage{multicol}

\usepackage{pifont}
\usepackage[ruled,vlined]{algorithm2e}
\newcommand{\cmark}{\ding{51}}%
\newcommand{\xmark}{\ding{55}}%

\usepackage[pagebackref,breaklinks,colorlinks]{hyperref}

\usepackage[capitalize]{cleveref}
\crefname{section}{Sec.}{Secs.}
\Crefname{section}{Section}{Sections}
\Crefname{table}{Table}{Tables}
\crefname{table}{Tab.}{Tabs.}

\def\cvprPaperID{2646} 
\def\confName{CVPR}
\def\confYear{2022}

\begin{document}

\title{Make A Long Image Short: Adaptive Token Length for Vision Transformers}

\renewcommand{\thefootnote}{\fnsymbol{footnote}}
\author{Yichen Zhu$^{1}$, Yuqin Zhu$^{1}$, Jie Du$^{1,2}$, Yi Wang$^{1}$, Zhicai Ou$^{1}$, Feifei Feng$^{1}$ and Jian Tang$^{1}$\footnotemark[4]\\
$^{1}$Midea Group, AI Innovation Center \\
$^{2}$Tongji University
}
\maketitle
\footnotetext[4]{Corresponding author.}
\renewcommand{\thefootnote}{\arabic{footnote}}

\begin{abstract}
The vision transformer splits each image into a sequence of tokens with fixed length and processes the tokens in the same way as words in natural language processing. More tokens normally lead to better performance but considerably increased computational cost. Motivated by the proverb "A picture is worth a thousand words" we aim to accelerate the ViT model by making a long image short. To this end, we propose a novel approach to assign token length adaptively during inference. Specifically, we first train a ViT model, called Resizable-ViT (ReViT), that can process any given input with diverse token lengths. Then, we retrieve the "token-length label" from ReViT and use it to train a lightweight Token-Length Assigner (TLA). The token-length labels are the smallest number of tokens to split an image that the ReViT can make the correct prediction, and TLA is learned to allocate the optimal token length based on these labels. The TLA enables the ReViT to process the image with the minimum sufficient number of tokens during inference. Thus, the inference speed is boosted by reducing the token numbers in the ViT model. Our approach is general and compatible with modern vision transformer architectures and can significantly reduce computational expanse. We verified the effectiveness of our methods on multiple representative ViT models (DeiT~\cite{deit}, LV-ViT~\cite{lvvit}, and TimesFormer~\cite{bertasius2021space}) across two tasks (image classification and action recognition).

\end{abstract}

\section{Introduction}
The vision transformer has achieved stunning success in computer vision since ViT~\cite{dosovitskiy2020image}. It has shown impressive capability upon convolutional neural networks (CNNs) on prevalent visual domains, including image classification~\cite{deit, twins}, object detection~\cite{detr, zhu2020deformable}, semantic segmentation~\cite{liu2021swin}, action recognition~\cite{mvit, bertasius2021space} with both supervised and self-supervised~\cite{he2021masked, bao2021beit} training configurations. Along with the development of ViT models, the deployment of vision transformers is becoming an issue due to its high computational cost. 
\begin{figure}[t]
    \begin{minipage}{1.0\linewidth}
    \centering
    \includegraphics[width =0.9\textwidth]{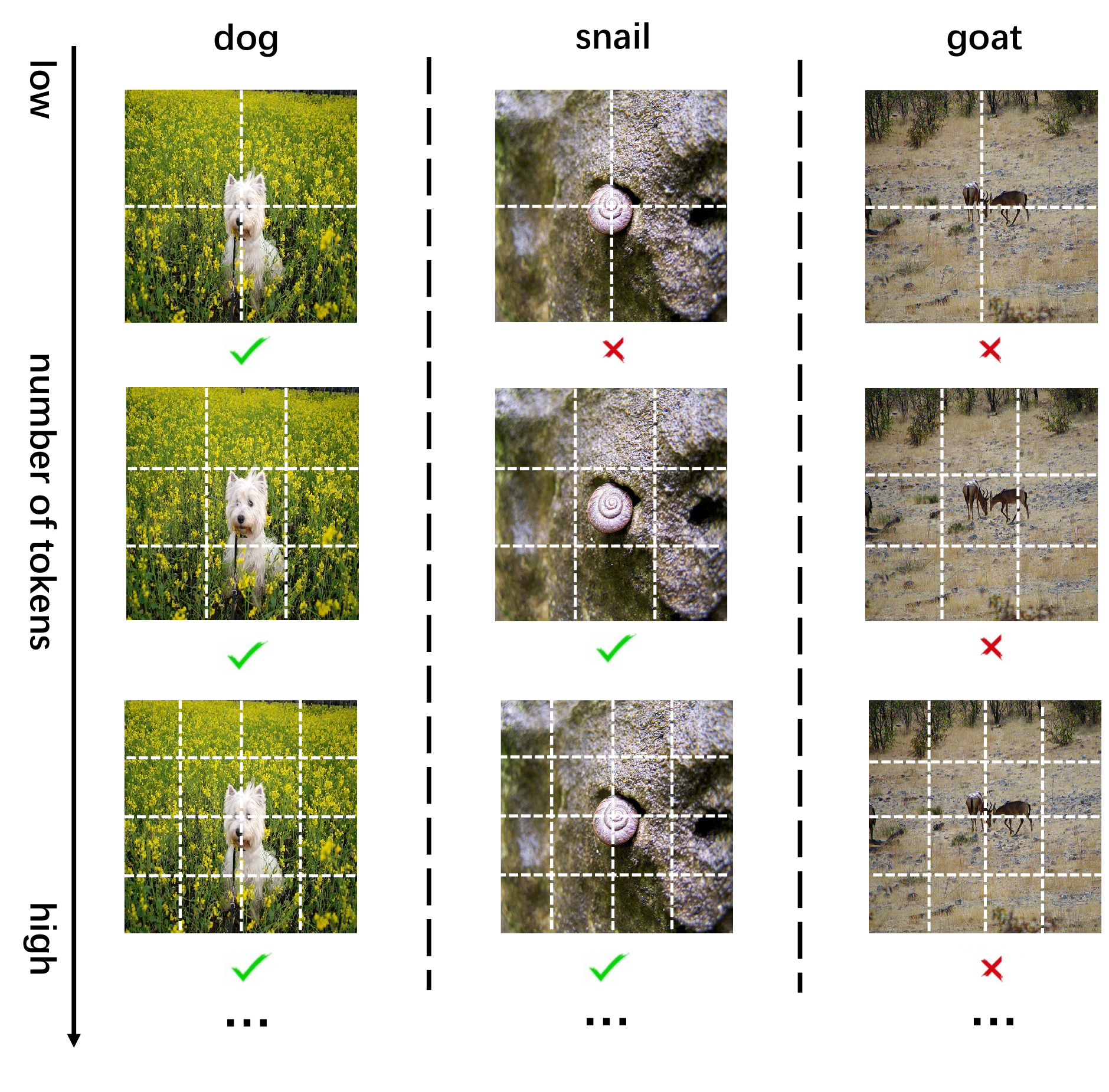}\\
    \end{minipage}
    \vspace{-3 mm}
      \caption{The motivation for our approach. While some images (right) may need many tokens to predict their category, some images are easy to recognize. Thus, only a small number of tokens is sufficient to classify them correctly.}\label{fig:motivation}
\end{figure}

How to accelerate ViT is an important yet rarely explored topic. Due to the substantial model difference between CNNs and ViT~\cite{naseer2021intriguing, mahmood2021robustness, raghu2021vision}, many techniques (i.e., pruning, distillation, and neural architecture search) to accelerate the CNNs cannot directly apply to ViT. As the attention module in the transformer computes the fully-connected relations among all of the input patches~\cite{vaswani2017attention}, the computational cost is then quadratic with regard to the length of the input sequence~\cite{choromanski2020rethinking, beltagy2020longformer}. As a result, the transformer suffers heavy computational costs, especially when the input sequence is long. In the ViT model, the images are split into a fixed number of tokens. Following the conventional paradigm~\cite{dosovitskiy2020image}, an image is represented by $16 \times 16$ tokens. We target to reduce the computational complexity of ViT from the perspective of reducing the number of tokens that are used to split the images. Our motivation is illustrated in Figure~\ref{fig:motivation}, which shows three examples predicted by the ViT model with three different token lengths. The results are obtained from three individually trained DeiT-S~\cite{deit} with three token lengths, while the check-mark denotes correct prediction and cross denotes the wrong prediction. We can observe that some "easy-to-classify" images only need 2$\times$2 tokens to correctly determine their category (i.e., the dog image on the left). In contrast, some images might need much more tokens to make the right prediction. These observations motivate us that the computational complexity of the existing ViT model can be drastically reduced if the input can be accurately classified with the minimum number of tokens.

Ideally, suppose we know the minimum number of tokens that correctly predict an individual image. In that case, a descent model can be trained with this information to assign the "optimal" token lengths for the ViT model. Therefore, we decouple the task into the ViT training and token length assignment. A naive approach is to train multiple ViT models, each dealing with one token length. However, it would be computational prohibitive in both training and testing. To tackle this issue, we begin with modifying a transformer from "static," representing that the model can only evaluate on a fixed, single token length, to "dynamic," where the ViT model can adaptively process the images with multiple token lengths. This dynamic vision transformer, called Resizable-ViT (ReViT), can mark the minimum sufficient token length used in the ViT model to obtain the correct prediction for each image. Then, we train a lightweight Token-Length Assigner (TLA) to predict an appropriate token length for the given image, where the label is retrieved from the ReVit. As such, the ReViT can use significantly less computational cost based on the new token length to split the images.

The biggest challenge of this approach is how to train the ReViT so that the ViT model can manage to process the image with any given size provided by the TLA. We introduce a token length-aware layer normalization to switch the normalization statics for each type of token length and a self-distillation module to improve the model performance of using short token length in ReViT. Another issue is that the ViT mode has to "see" the images with corresponding token lengths beforehand to make it capable of dealing with the various token lengths for images. As the number of predefined token-length choices increases, the training cost increase linearly. To reduce the training cost, we introduce a parallel computing strategy for efficient training, which enables training of ReViT is almost as cheap as training a vanilla ViT model. 

We illustrate the effectiveness of our method on representative ViT models, including DeiT~\cite{deit} and LV-ViT~\cite{lvvit} on image classification, as well as TimesFormer~\cite{bertasius2021space} on video recognition. Our experiments demonstrate that our approach can effectively reduce the computation cost while maintaining performance. For example, we manage to speed up the DeiT-S~\cite{deit} model by 50\% with accuracy drop by 0.3\%. On action recognition, the computational cost of TimesFormer~\cite{bertasius2021space} is reduced up to 33\% on Kinetic 400 while only sacrificing 0.5\% recognition accuracy. 

\begin{figure*}[!ht]
    \begin{minipage}{1.0\linewidth}
    \centering
    \includegraphics[width =0.8\textwidth]{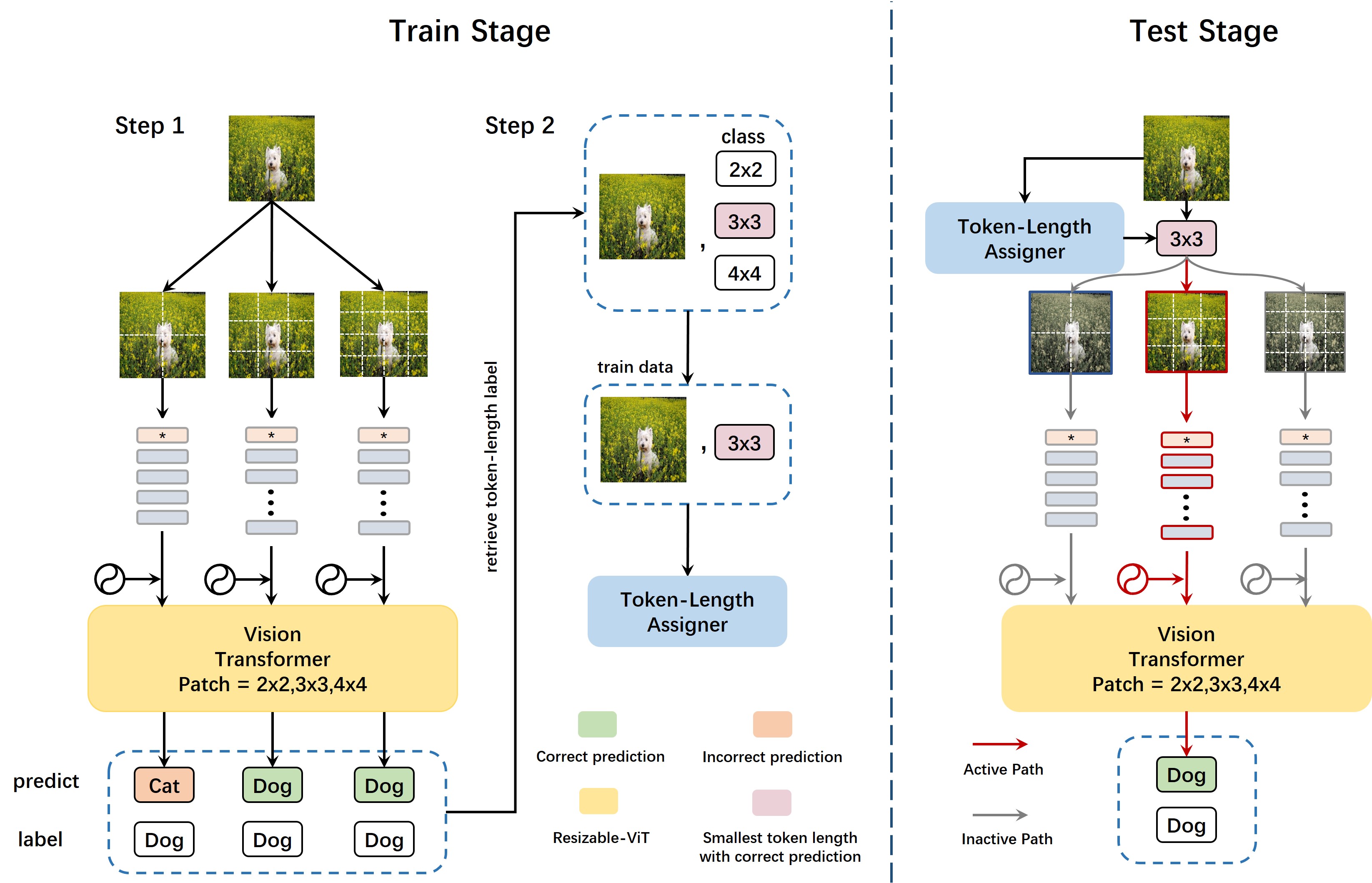}\\
    \end{minipage}
    \vspace{-3 mm}
      \caption{\textbf{Left}: There are two steps in the training procedure. First, we train the Resizable-ViT that can split an image into any predefined token length. Secondly, we train a Token-Length Assigner based on the token-length label that is retrieved from ReViT. It is the smallest number of tokens that can correctly predicate the class of the image. \textbf{Right}: In inference, the TLA first assigns a token-length for the image, then ReViT uses this setting to make predication.}\label{fig:overview}
\end{figure*}

\section{Related Works}
\noindent
\textbf{Vision Transformer.} Transformers~\cite{vaswani2017attention} have drawn much attention to computer vision recently due to
its strong capability of modeling long-range relation. Many attempts have been made to integrate the long-range modeling into CNNs, such as non-local networks~\cite{wang2018non, yin2020disentangled}, relation networks~\cite{hu2018relation}, etc. Vision Transformer (ViT)~\cite{dosovitskiy2020image} first introduced a set of pure Transformer backbones for image classification, and its follow-ups modify the vision transformer soon to dominate many downstream tasks for computer vision, such as object detection~\cite{carion2020end, zhu2020deformable}, semantic segmentation~\cite{liu2021swin}, action recognition~\cite{bertasius2021space, mvit}, 2D/3D human pose estimation~\cite{yang2020transpose, poseformer}, and 3D object detection~\cite{pointformer}. It has shown great potential to be an alternative backbone for convolutional neural networks. 
\\
\\
\noindent
\textbf{Model Compression.} It is known that the over-parameterized model has many attracting merits and can achieve better performance than small models. However, computational efficiency is critical in real-world scenarios, where the executed computation is translated into power consumption or carbon emission. Many works have tried on reducing the computational cost of CNNs via neural architecture search~\cite{liu2018darts, zoph2016neural, fbnnas, chu2021fairnas, guo2020single}, knowledge distillation~\cite{hinton2015distilling, zhu2021student}, dynamic routing~\cite{elbayad2019depth, cai2019once, wang2020glance, zhu2019resizable, yu2018slimmable} and pruning~\cite{han2015deep, frankle2018lottery}, but how to accelerate the ViT model have been rarely explored.

Recently, some attempts have been made to compress the vision transformer model, including quantization~\cite{liu2021post}, distillation~\cite{li2021mst}, and pruning~\cite{tang2021patch}. Some works have started to reduce the computational cost of the ViT models in terms of the token length. These works can be classified into two forms, unstructured token sparsification and structure token division. Most works, including PatchSlim~\cite{tang2021patch}, TokenSparse~\cite{rao2021dynamicvit}, GlobalEncoder~\cite{song2021dynamic},  IA-RED~\cite{pan2021ia}, and Tokenlearner~\cite{ryoo2021tokenlearner}, focus on the former. They aim to remove the uninformative tokens, such as tokens that learn features from the background of the image. As a result, the inference speed is boosted by reserving the informative tokens only. These approaches typically need to progressively reduce the number of tokens based on the inputs and can be performed either jointly with ViT training or afterward. 
The latter, unstructured token sparsification, is the most related work to us. Wang et al.~\cite{wang2021not} proposed DVT to determine the number of patches to divide an image dynamically. Specifically, they leverage a cascade ViT models, where each ViT is responsible for one type of token length. The cascade ViT makes a sequential decision. It will stop inference for an input image if it has sufficient confidence in the prediction on the current token length. Different from DVT~\cite{wang2021not}, our method is more accessible and practical since only a \textit{single} ViT model is required. Moreover, we pay more attention on how to \textit{accurately} decide the smallest number of token lengths that can give correct predication in the transformer for each image.
\begin{figure*}[!ht]
    \begin{minipage}{1.0\linewidth}
    \centering
    \includegraphics[width =0.65\textwidth]{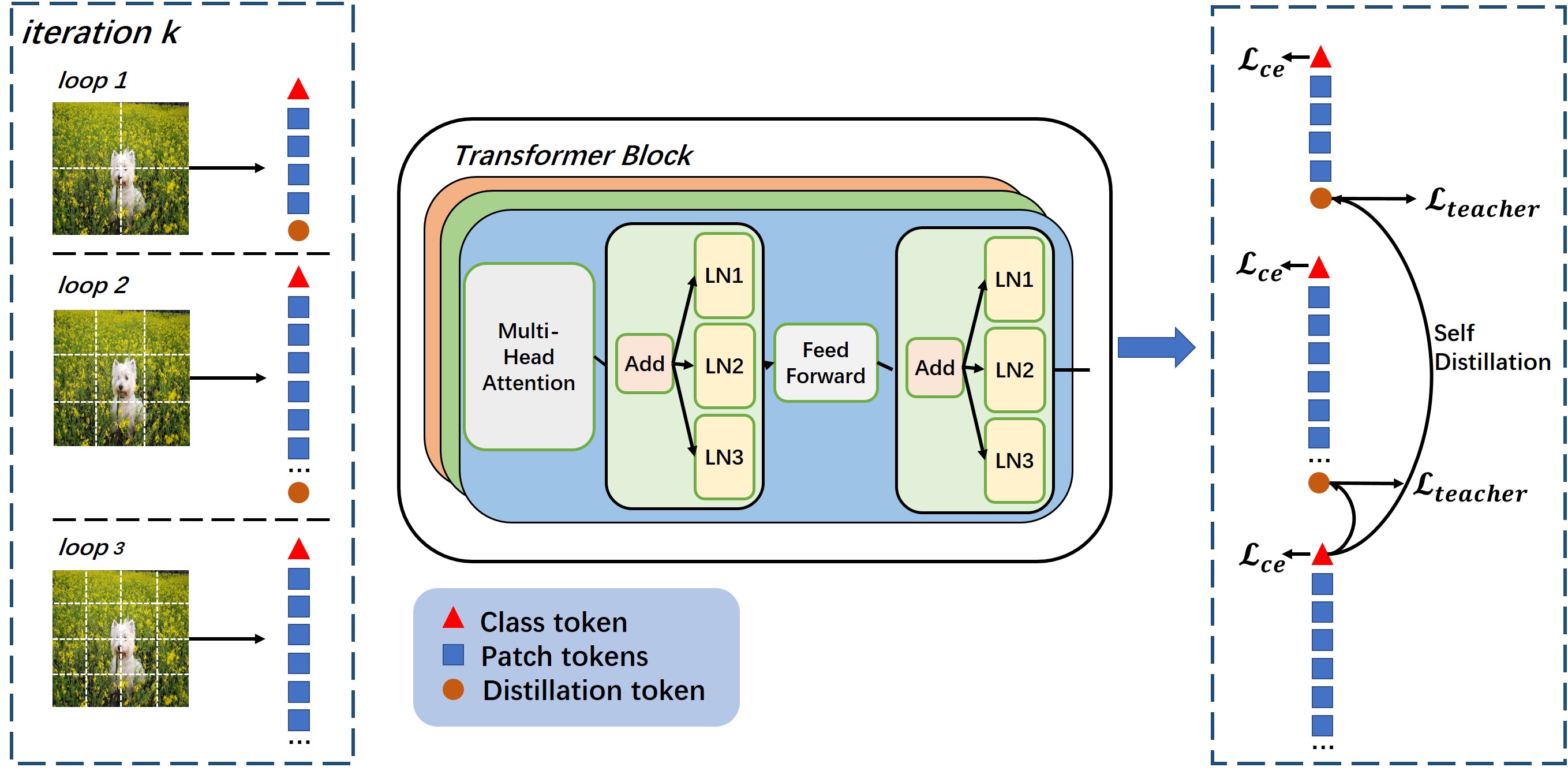}\\
    \end{minipage}
    \vspace{-3 mm}
      \caption{Example of self-distillation and token-length aware layer normalization in ReViT. Each token length corresponds to a LayerNorm (LN in this figure) and pass-through this LayerNorm during both training and inference. The self-distillation is only conducted in training, where smaller token lengths have an extra distillation token to learn from the teacher's knowledge.}\label{fig:betterstudetn}
\end{figure*}
\section{Methodology}
The vision transformers consider an image as a sentence. It splits each 2D image into the 1D tokens and models the long-range dependency between tokens with the multi-head self-attention mechanism. The self-attention has been recognized as the computation bottleneck in the transformer model. Its computational cost increases quadratically to the number of incoming tokens. As aforementioned, our approach is motivated by the fact that many "easy-to-recognize" images do not require $16\times16$ tokens~\cite{dosovitskiy2020image} to correctly predication their category. Thus, the computational cost can be saved by processing fewer tokens on "easy" images while using more tokens on "hard" images. It is worth noting that the key to a successful input-dependent token-adaptive ViT model is to know precisely the minimum number of tokens that are sufficient to correctly classify the image.

Therefore, we decouple the model training into two stages. At the first stage, we train a ViT model that can process an image with any predefined token lengths. Normally, a single ViT model can only deal with one token length. We discuss the detailed model design and training strategy of such model in Section~\ref{sec:revit}. We need to train a model to assign the token length to the images at the second stage. First, we obtain the token-length label, the minimum number of tokens that this ViT model needs to perform correct classification, from the previously trained ViT model. Then, a Token-Length Assigner (TLA) is trained with the training data, where the input is an image and the label is the token length. Intuitively, such decoupled procedure helps the TLA to make a better decision on selecting the token lengths. In inference, the TLA directly guides the ViT model that how many tokens are sufficient to make the decision based on the input. The complete training and testing process are shown in Figure~\ref{fig:overview}.

In the following, we first introduce the Token-Label Assigner, then present the training method on the Resizable-ViT model and improved techniques.

\subsection{Token-Length Assigner}
The purpose of the Token-Length Assigner (TLA) is to make precise predictions based on the feedback from ReViT. The training of TLA is performed after the ReViT. We first define a list of token-length $L = [l_{1}, l_{2}, \dots, l_{n}]$, where the token length is in descending order. For notation simplicity, we use a single number to denote the token length. For example, $L = [14 \times 14, 10 \times 10, 7 \times 7]$. The model with tokens of $7 \times 7$ has the lowest computational cost among the three token lengths.

To train a TLA, we need to retrieve a token-length label from the trained ReViT. We define the token-length label for an image as the smallest token length that the ViT model needs to correctly classify this particular image. For instance, the inference speed of the ReViT $M$ would be $Speed(M_{l_{1}}) < Speed(M_{l_{2}}) < \dots < Speed(M_{l_{k}})$. The $k = len(L)$ denotes the total number of choices for token length. We can obtain the predication of $y$ for each input $x$, $y_{l_{i}} = M_{l_{i}}(X), \forall i \in n$. The label of the input $x$ is the smallest token size $l_{j}$ that for any smaller token length cannot make the correct prediction $y_{l_{j-1}} \neq y^{gt}$, the $gt$ is the ground truth label. As such, we would have a set of input-output $(x, l_{j})$, which is used to train the TLA. The TLA is a lightweight module since the token-label assignment is an easy task with only a few labels. The additional computational overhead introduced by this TLA module is quite small, especially considering the computational overhead saved by reducing the unnecessary tokens in the ViT model.

\subsection{Resizable-ViT} \label{sec:revit}
This section introduces the Resizable-ViT (ReViT), a dynamic ViT model that can predict the category of a given image with various token lengths. We introduce two techniques that improve the performance of ReViT, then present the training strategy along with an efficient training implementation to accelerate the training of ReViT.
\\
\\
\noindent
\textbf{Token-Aware Layer Normalization.}
Layer Normalization (LN/LayerNorm) layer is a standard normalization method to accelerate training and improve the generalization of the Transformer architecture. It is common for both natural language processing and computer vision to adopt an LN after addition in transformer block. Because the feature maps of both self-attention matrices and feed-forward networks are constantly changing, as the number of token size changes during training, it leads to inaccurate normalization statistics across the different token lengths in shared layers, which impairs the test accuracy. We also empirically find that the LN cannot share in ReViT.

To tackle this issue, we propose a Token-Length-Aware LayerNorm (TAL-LN), which uses an independent LayerNorm corresponding to each choice of token length in the predefined token length list. In other words, we use a ”$Add$ $\&$ $\{LN_{1}$, ..., $LN_{k}\}$” as a building block, where k denotes as the number of the predefined token length. As such, each LayerNorm layer specifically calculates layer-wise statistics and learns the parameters of the corresponding feature map. Moreover, the number of extra parameters in TAL-LN is negligible since the number of parameters in normalization layers usually takes less than one percent of the total model size~\cite{yu2018slimmable}. A brief summary is shown in Figure~\ref{fig:betterstudetn}.
\\
\\
\noindent
\textbf{Self-Distillation}
It is aware that the performance of ViT is strongly correlated to the number of patches, and experiments show that reducing the token size significantly hampers the accuracy of small token ViT. Directly optimizing via the supervision from the ground truth poses difficulty for the small token length sub-model. Motivated by self-attention, a variant of knowledge distillation techniques, where the teacher can be insufficiently trained or even the student model itself~\cite{yu2018slimmable, zhu2021student, yu2019universally, yu2020bignas}, we present a token length-aware self-distillation (TLSD). We will show in the next section, the model with the largest token length $M_{1}$ is always trained first. For $M_{l_{1}}$, the training objective is to minimize the cross-entropy loss $\mathcal{L}_{CE}$. When comes to the model with other token length $M_{l_{i}}, i \leq k, i \neq 1$, we use a distillation objective to train the target model:
\begin{equation}
    \mathcal{L}_{teacher} = (1 - \lambda)\mathcal{L}_{CE}(\phi(Z_{s}), y) + \lambda \tau^2KL(\phi(Z_{s}/\tau), \phi(Z_{t}/\tau))
\end{equation}

where $Z_{s}$ and $Z_{t}$ is the logits of the student model $M_{l_{i}}$ and teacher model $M_{l_{1}}$, respectively. $\tau$ is the temperature for the distillation, $\lambda$ is the coefficient balancing the KL loss (Kullack-Leibler divergence) and the CE loss (cross-entropy) on ground truth label $y$, and $\phi$ is the softmax function. Similar to DeiT, we add a distillation token for student models. Figure~\ref{fig:betterstudetn} gives an overview. Notably, this distillation scheme is computational-free: we can directly use the predicted label of the model with the largest token length as the training label for other sub-model, while for the largest token length model, we use ground truth.

\begin{figure}[t]
    \begin{minipage}{1.0\linewidth}
    \centering
    \includegraphics[width =0.7\textwidth]{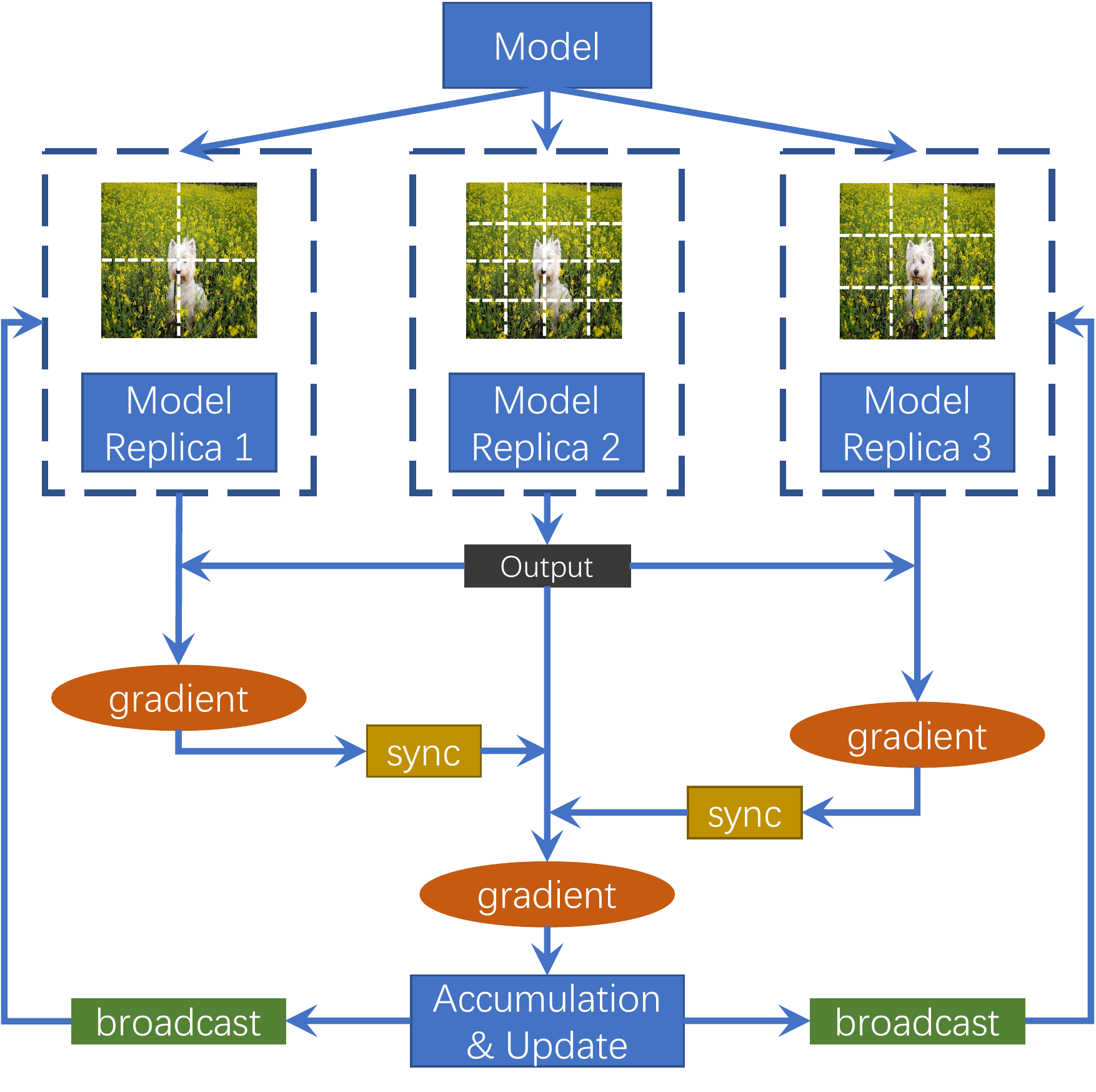}\\
    \end{minipage}
    \vspace{-3 mm}
      \caption{Efficient training implement for Resizable Transformer through parallel computing. All gradient from the replicate nodes are synchronize on the node where that have the largest token length to save the cost of communication.}\label{fig:parallel}
\end{figure}

\subsection{Training Strategy}
In order to let the ViT model adaptively process the token length in the predefined choice list, an image with various token lengths has to be "seen." Inspired by the batch gradient accumulation, a technique to bypass the issue of small batch size by accumulating the gradient and batch statistics in a single iteration (updating the model parameters counts for one iteration), we propose a mixing token length training. As illustrated in Algorithm~\ref{alg:alg1}, a batch of images that are processed with different token lengths are calculated the loss through in the feed-forward, and then obtain the individual gradients. After all choices of token length are looped over, the gradients of all parameters calculated by feeding different token lengths are accumulated to perform parameter updating. 
\begin{algorithm}[t]
\label{alg:alg1}
\SetAlgoLined
\textbf{Require:} Define Token-Length Assigner $T$, token-length list $\mathbf{R}$, for example, $\{16, 24, 32\}$. The iterations $N_M$ for training $M$. The $CE(\cdot)$ denotes cross-entropy loss, and $DisT(\cdot)$ denotes distillation loss.\\
\For{\upshape $t = 1, \dots, N_{M}$} {
    \upshape
    Get data $x$ and class label $y_{c}$ of current mini-batch.\\
    Clear gradients for all parameters, \textit{optimizer.zero\_grad()}\\
    \For{\upshape $i = 1, \dots, len(\mathbf{R}) - 1$}
    {
    Convert ReviT to selected token-length $M_{i}$,  \\
    Execute current scaling configuration. $\hat{y}_{i} = M_{i}(x)$.\\
    \If{$\mathbf{R}[i]$ == 16}{
      set teacher label. $\hat{y}_{i}^{teacher} =\hat{y}_{i}$ \\
      Compute loss \textit{$loss_{i} = CE(\hat{y}_{i}, y)$} \\
    }\Else{
      Compute loss \textit{$loss_{i} = DisT(\hat{y}_{i}^{teacher}, \hat{y}_{i}, y)$} \\
    }
    Compute gradients, \textit{$loss_{i}.backward()$} \\
    }
    Update weights, \textit{$optimizer.step()$.} \\
    }
Obtain token-length label for all train data $(x, y_{t})$.\\
Train $T$ with $(x, y_{t})$.\\
\caption{Training Resizable-ViT $M$.}
\label{alg:alg1}
\end{algorithm}
\\
\\
\noindent
\textbf{Efficient Training Implementation.} 
An issue with the above training strategy is that the training time is linearly increasing as the number of the predefined number of token lengths choices increases. Here, we propose an efficient implementation strategy to trade memory cost to training cost. As illustrated by Figure~\ref{fig:parallel}, we replicate the model, each model corresponds to a token length. At the end of each iteration, 
\begin{figure*}[!ht]
    \begin{minipage}{1.0\linewidth}
    \centering
    \includegraphics[width =0.8\textwidth]{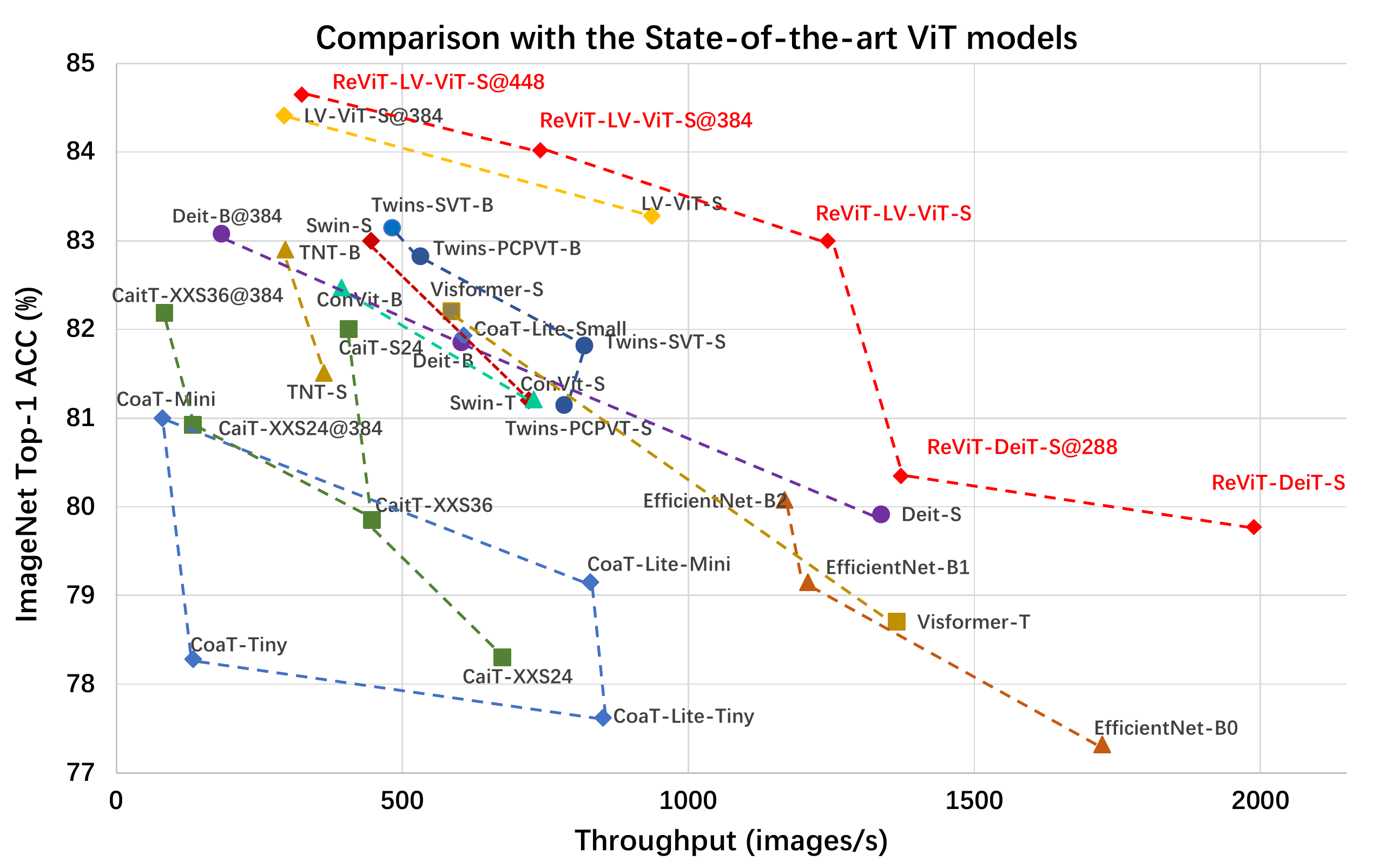}\\
    \end{minipage}
    \vspace{-3 mm}
      \caption{Comparison of different models with various accuracy-throughput trade-off. The throughput is measured on an NVIDIA RTX 3090 GPU with batch size fixed to 32. The input image size is $224 \times 224$ unless indicate otherwise. The ReViT (red square in the figure) achieves better trade-off than other methods. }\label{fig:overall_perm}
\end{figure*}

the gradients of different replicas are0 synchronized and accumulated. We always send the gradient of replicas in which the token length is small to the one with a larger token length because they are the training bottleneck. Thus the communication cost in the gradient synchronization step is free. Then the parameters in the model are updated through back-propagation. After updating the parameters is completed, the main process distributes the learned parameters to the rest of the replicas. These steps are repeated until the end of the training, and all replicas except the model in the main process can be dislodged. As such, the training time of the Resizable Transformer reduces from $O(k)$ to $O(1)$, where $k$ is the number of the predefined token length. Though the number of $k$ is small, i.e., $k = 3$, in practice, the computation of $k$ itself is high. Through our designed parallel computing, training the Resizable Transformer is almost the same as the naive ViT, where the cost from communication between replicas is negligible compared to the model training cost. In exchange for fast training, extra computational power is required for parallel computing.

\section{Experiments}
\noindent
\textbf{Implementation details.} For ImageNet classification, We train all of the models on the ImageNet~\cite{deng2009imagenet} training set with approximately 1.2 million images and report the accuracy on the 50k images in the test set. By default, the predefined token length is set to $14 \times 14$ and $7 \times 7$. We didn't use token of of 4 $\times$ 4 as shown in the motivation figure, since the accuracy drop significantly with this number of tokens. We conduct experiments on DeiT-S~\cite{deit} and LV-ViT-S~\cite{lvvit} with image resolution in training and testing is 224 × 224 unless otherwise specified. For the training settings and optimization methods, we simply follow those in the original papers of DeiT~\cite{deit} and LV-ViT~\cite{lvvit}. For LV-ViT, we obtain the token label for smaller token length according to their method. Since our method is generally applicable to diverse ViT models, We didn't carefully finetune the hyper-parameters. We also train the ReViT on resized images with higher resolutions, 228 on DeiT-S and 448 on LV-ViT-S. Because the the convolutional layers with large kernel and stride are required to perform patch embedding, which may cause difficulty on optimization. Thus, we replace the large kernel convolution with consecutive convolutions followed Xiao et al.~\cite{xiao2021early}. After the ReViT finishes training, we obtain the token-length labels for all training data and train the Token-Length Assigner. The TLA is a shrink version of EfficientNet-B0. It is extremely small compared to the ViT model. We give detailed description of the TLA  architecture and training details in the Appendix.

We use Kinetics400~\cite{kay2017kinetics} to conduct experiments on action recognition. The Kinetics400 is a widely-used benchmark for action recognition, which includes around 240k training videos and 20k validation videos in 400 classes. We follow the training setting of TimeSformer~\cite{bertasius2021space}. Specifically, two versions of TimeSformer are tested. The TimeSformer, which is the default version of TimeSformer model operating on $8 \times 224 \times 224$ video clips, and TimeSformer-HR, a high
spatial resolution variant that operates on $16 \times 448 \times 448$ video clips. All models are pretrained on the ImageNet1K. The TimeSformer use default of patch size of $16 \times 16$, while we use the same setting as in ImageNet classification. 
\begin{figure*}[t]
    \centering
    \includegraphics[width =0.9\textwidth]{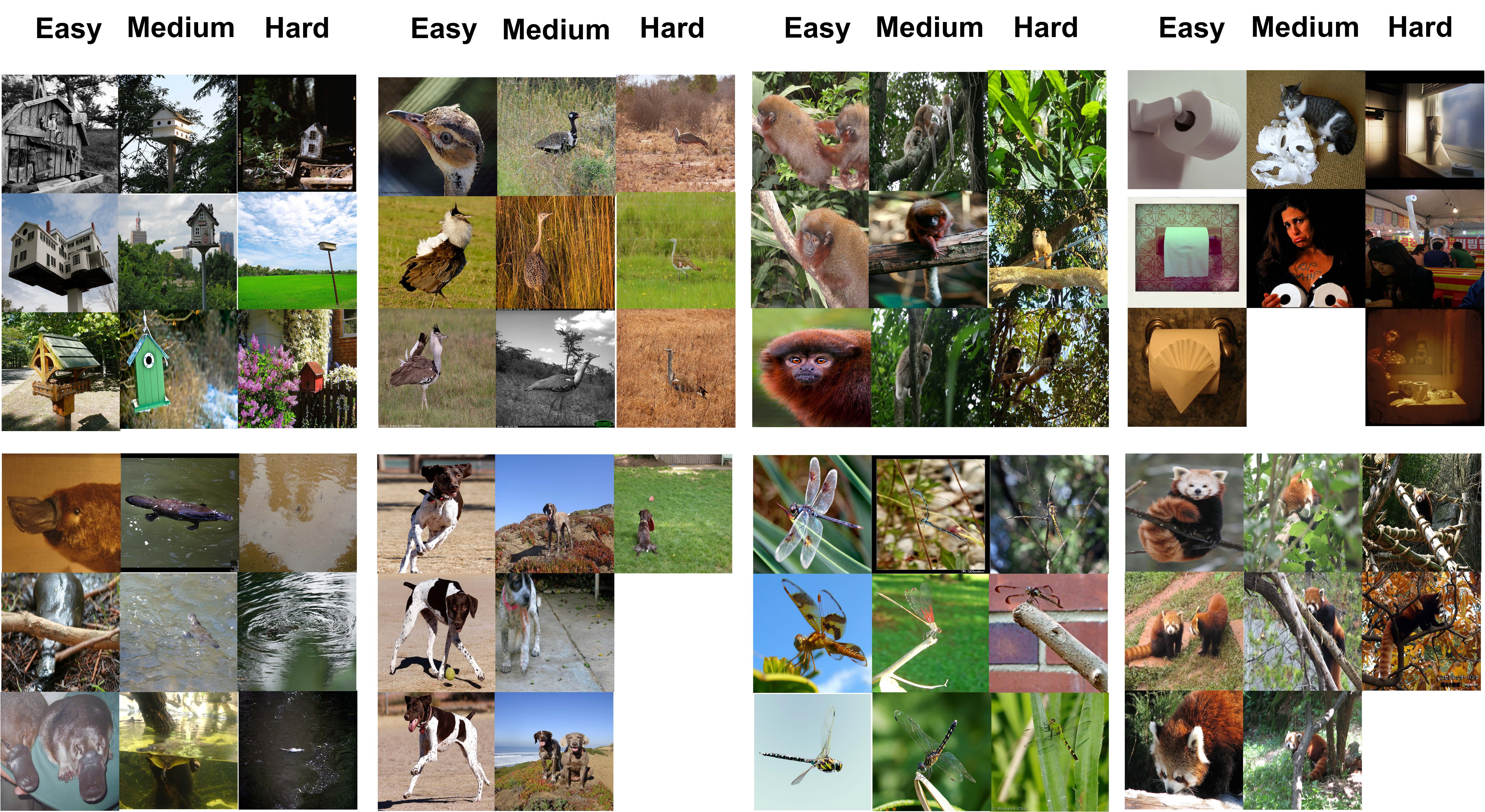}
    \vspace{-3 mm}
      \caption{Visualization of "hard", "medium", and "easy" samples that predicted by Token-Length Assigner and which the ReViT-DeiT-S got correction prediction. Most of the "easy" images have clear sight on the object, while size of objects is mostly small for "hard" samples.}\label{fig:visualization}
\end{figure*}
\\
\\
\noindent
\textbf{Main Results on ImageNet Classification.} We report the main result of our ReViT based on DeiT-S and LV-ViT-S in Figure~\ref{fig:overall_perm}. We compare our approach with a number of models, including DeiT~\cite{deit}, CaiT~\cite{cait}, LV-ViT~\cite{lvvit}, CoaT~\cite{coat}, Swin~\cite{liu2021swin}, Twins~\cite{twins}, Visformer~\cite{visformer}, ConViT~\cite{wu2021cvt}, TNT~\cite{tnt}, and EfficientNet~\cite{tan2019efficientnet}. It shows that our method can achieve a plausible accuracy-throughput trade-off. Specifically, the ReViT reduces the computational cost of the baseline counterpart by reducing the token number used for inference. For instance, the inference speed of DeiT-S is increased by 50\% with a 0.3\% accuracy drop. By increasing the input resolution, we manage to outperform the baseline counterpart given a similar computational cost. The ReViT based on LV-ViT-S using $448 \times 448$ image size achieves 0.2\% higher top-1 accuracy with slightly faster inference speed than LV-ViT-S with $384 \times 384$.
\\
\\
\noindent
\textbf{Main Results on Kinetic 400 Action Recognition.} We further apply ReviT to the video action recognition. We train the ReviT-TimeSformer and ReviT-TimeSformer-HR and compare our method with the baseline TimeSformer and TimeSformer-HR, respectively. We list the results in Table~\ref{tbl:timesformer}. We can see that our method speed up the TimeSformer baseline. For TimeSformer, we reduce the computational cost of the baseline by approximately 33\% with 0.5\% accuracy drop. By training on larger image resolution, we correspondingly reduce the model by 28\% with 0.4\% accuracy drop. It is slightly worse than smaller resolution. Nevertheless, our experiments show that ReViT works generally well on action recognition. 
\begin{table}
\centering
\begin{tabular}{l|c|c|c}
\toprule
Model & Pretrain & Acc.(\%) & TFLOPs \\
\midrule
TimeSformer & IN-1K & 75.8 & 0.59 \\
ReViT-TimeSformer & IN-1K & 75.3 & 0.39\\
\midrule
TimeSformer-HR & IN-1K & 77.8 &  5.11  \\
ReViT-TimeSformer-HR & IN-1K & 77.4 & 3.68\\
\midrule
\bottomrule
\end{tabular}
\caption{Comparing ReViT to TimeSformer. The ReViT is built based on TimeSformer. All methods are pretrained with ImageNet1K (IN-1K in table).}
\label{tbl:timesformer}
\end{table}
\\
\\
\noindent
\textbf{Visualization of samples with different token-length.} \label{sec:visual} We select eight classes from the ImageNet validation set and pick three samples from three categories, easy, medium, and hard, corresponding to tokens with $14 \times 14$, $10 \times 10$, and $7 \times 7$, respectively. It is selected according to the token length assigned by the Token-Length Assigner. The images are shown in Figure~\ref{fig:visualization}. Note that some classes do not have all images filled in because less than three samples in the validation set belong to this category. For instance, only one image in the dog class need to use the largest token length to do the classification. We observe that the number of token lengths needed to predict their classes is correlated to the object's size. If the object is large, only a few tokens are sufficient to predicate their category. 

\subsection{Ablation Study}
\noindent
\textbf{Shared patch embedding and position encoding.}
We experiment to see what is the impact of using share patch embedding and position encoding. Because the token number changes during training, we apply some techniques in order to share both operations. For position encoding, we follow the ViT~\cite{dosovitskiy2020image} to zero-pad the position encoding module whenever the token size changes. This technique is used initially to adjust the positional encoding in the pretrain-finetune paradigm. For shared patch embedding, we use weight-sharing kernel~\cite{cai2019once}. A large kernel is constructed to process a large patch size. When the patch size changes, a smaller kernel with shared weight on the center is adopted to flat the image patch. 

As shown in Table~\ref{tbl:sharep}, both shared patch embedding and shared positional encoding decrease the model's accuracy. Especially for the sharing patch strategy, the accuracy dropped nearly 14\% on the large token-lengths' model. The share positional encoding module performs better than shared patch embedding, but still severally hurt the performance of ReViT. 
\begin{table}
\centering
\begin{tabular}{l|c|c|c|c|c}
\toprule
\multirow{2}{*}{Method} & \multirow{2}{*}{SD*} & \multirow{2}{*}{$\tau$}  & \multicolumn{3}{c}{Top-1 Acc (\%)} \\
   &       &      &     14 $\times$ 14 & 10 $\times$ 10 & 7 $\times$ 7 \\
\midrule
Deit-S  & \xmark    &   -    &  79.85 & 74.68  & 72.41  \\
\midrule
\multirow{3}{*}{ReViT} &   \xmark  &  -  & 80.12 & 74.24 & 70.15\\
& \cmark & 0.5 &  79.92 & 76.16  & 71.33 \\
& \cmark & 0.9 & 79.83  &  76.86 & 74.21 \\
\bottomrule
\end{tabular}
\caption{The ablation study of self-distillation in ReviT. The SD* denotes the self-distillation. We also evaluate the performance with different choices of $\tau$. The self-distillation improve the performance notably, the small token length model outperform the baseline when $\tau = 0.9$.}
\label{tbl:self_distill}
\end{table}
\begin{table}
\centering
\begin{tabular}{l|c|c|c|c|c}
\toprule
\multirow{2}{*}{Method} & \multicolumn{2}{c}{Shared}   & \multicolumn{3}{c}{Top-1 Acc (\%)} \\
   &    Patch   &   Pos   &     14 $\times$ 14 & 10 $\times$ 10 & 7 $\times$ 7 \\
\midrule
\multirow{3}{*}{ReViT} &   \xmark  &  \cmark  & 65.14 &  61.30 &  58.35\\
& \cmark & \xmark & 75.24 & 71.32  & 69.73 \\
& \cmark & \cmark & 79.83  &  76.85 & 74.21 \\
\bottomrule
\end{tabular}
\caption{The ablation study of shared patch embedding and shared position encoding in ReViT. The Pos denotes the positional encoding module. We notice that share these two modules decrease the model accuracy.}
\label{tbl:sharep}
\end{table}
\\
\\
\noindent
\textbf{The effect of self-distillation and choice of $\tau$.} \label{sec:selfdistill_exp} We conduct experiments to verify the effectiveness of self-distillation in ReViT and discuss the impact of hyper-parameters $\tau$. We test two different $\tau$, 0.9 and 0.5, for all sub-networks. We demonstrate the results in Table~\ref{tbl:self_distill}. Without self-distillation, the accuracy on small token length is comparable with tokens of $10 \times 10$, but much worse on tokens of $7 \times 7$. When applying self-distillation with $\tau = 5$, the accuracy of both models increases. The further evaluate the model use $\tau = 5$. The higher $\tau$ affects the accuracy of the largest token length, drops the accuracy around 0.3\%, but significantly increases the performance of $7 \times 7$ models. It shows the necessity of using self-distillation in our scenario.
\begin{figure}[!ht]
    \begin{minipage}{1.0\linewidth}
    \centering
    \includegraphics[height=4cm, width =\textwidth]{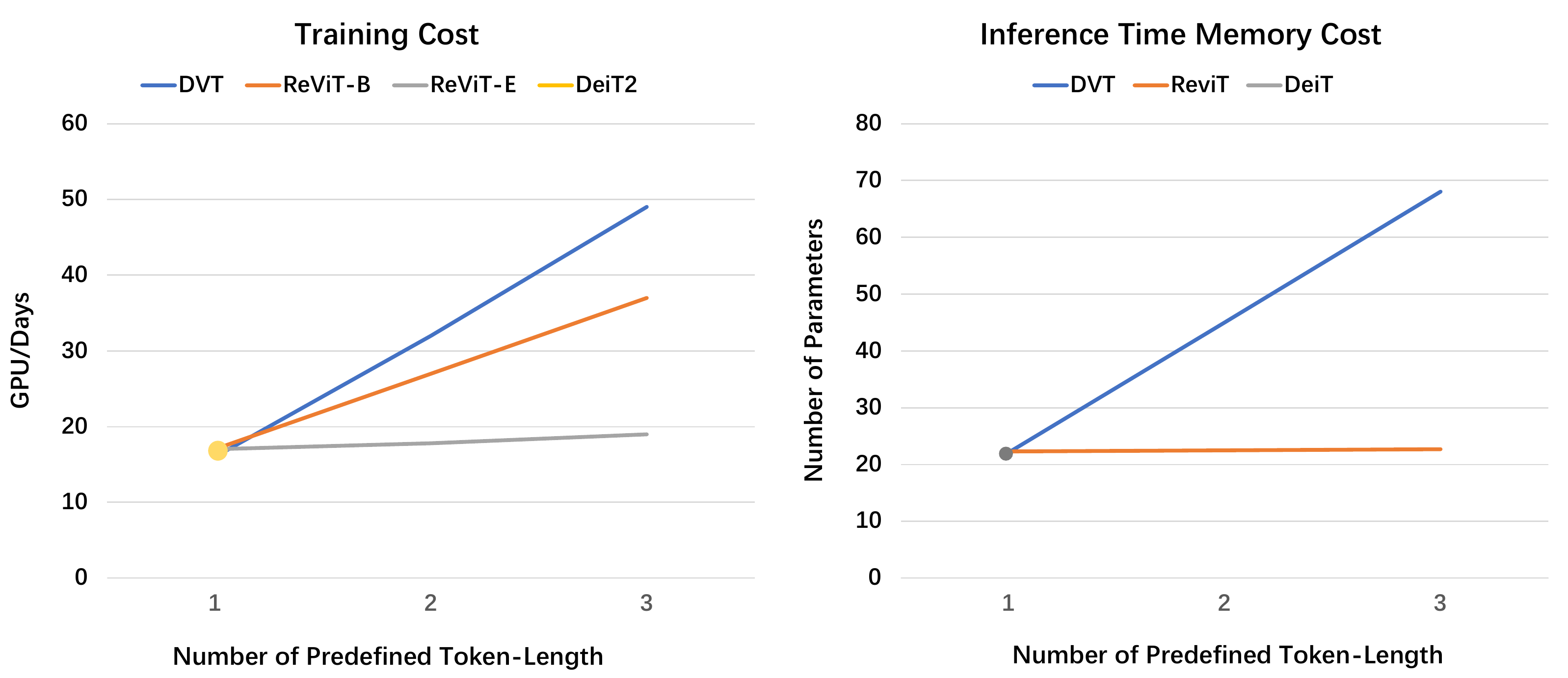}\\
    \end{minipage}
    \vspace{-3 mm}
      \caption{Compare our approach with DeiT-S~\cite{deit} and DVT~\cite{wang2021not} for training cost and memory cost at inference time in terms of the number of predefined token-length. Our proposed ReViT is almost a cheap as training the baseline DeiT-S, while DVT requires linearly increased budget on training and memory.}\label{fig:cos_ana}
\end{figure}
\\
\\
\textbf{Training cost and Memory Consumption.} \label{sec:traincost} We compare ReViT with DeiT-S and DVT~\cite{wang2021not} in terms of training cost and memory consumption, shown in Figure~\ref{fig:cos_ana}. The ReViT-B denotes the baseline approach of ReViT and ReViT-E is the efficient implementation method. We can observe that both ReViT-B and DeiT-S are increasing linearly as the choice in $s$ increases. The ReViT-B is cheaper because the backpropagation of multiple token lengths is merged. On the other hand, the training time of ReViT-E slightly increases due to the communication cost between parallel models increasing. 
For the memory consumption (number of parameters) during testing, since our method only has a single ViT where most computational heavy components are shared, the memory cost is slightly higher than the baseline. Compared to the DVT, the number of parameters increases with respect to the increasing number of token length choices is negligible. It indicates that our approach is more practical than DVT from the perspective of training cost and memory cost. Also, it is easier than DVT to apply our method on the existing ViT model. 

\section{Conclusion}
This paper aims to reduce the token length to split the image in ViT model to eliminate unnecessary computational costs. We propose the Resizable Transformer (ReViT), which can adaptively process any predefined token size for a given image. Then, we define a Token-Length Assigner to decide the minimum number of tokens that the transformer can use to classify the individual image correctly. Extensive experiments indicate that ReViT can significantly accelerate the state-of-the-art ViT model.

{\small
\bibliographystyle{ieee_fullname}
\bibliography{egbib}

\begin{thebibliography}{10}\itemsep=-1pt

\bibitem{bao2021beit}
Hangbo Bao, Li Dong, and Furu Wei.
\newblock Beit: Bert pre-training of image transformers.
\newblock {\em arXiv preprint arXiv:2106.08254}, 2021.

\bibitem{beltagy2020longformer}
Iz Beltagy, Matthew~E Peters, and Arman Cohan.
\newblock Longformer: The long-document transformer.
\newblock {\em arXiv preprint arXiv:2004.05150}, 2020.

\bibitem{bertasius2021space}
Gedas Bertasius, Heng Wang, and Lorenzo Torresani.
\newblock Is space-time attention all you need for video understanding?
\newblock {\em arXiv preprint arXiv:2102.05095}, 2021.

\bibitem{cai2019once}
Han Cai, Chuang Gan, Tianzhe Wang, Zhekai Zhang, and Song Han.
\newblock Once-for-all: Train one network and specialize it for efficient
  deployment.
\newblock {\em arXiv preprint arXiv:1908.09791}, 2019.

\bibitem{detr}
Nicolas Carion, Francisco Massa, Gabriel Synnaeve, Nicolas Usunier, Alexander
  Kirillov, and Sergey Zagoruyko.
\newblock End-to-end object detection with transformers.
\newblock In {\em European Conference on Computer Vision}, pages 213--229.
  Springer, 2020.

\bibitem{carion2020end}
Nicolas Carion, Francisco Massa, Gabriel Synnaeve, Nicolas Usunier, Alexander
  Kirillov, and Sergey Zagoruyko.
\newblock End-to-end object detection with transformers.
\newblock In {\em European Conference on Computer Vision}, pages 213--229.
  Springer, 2020.

\bibitem{visformer}
Zhengsu Chen, Lingxi Xie, Jianwei Niu, Xuefeng Liu, Longhui Wei, and Qi Tian.
\newblock Visformer: The vision-friendly transformer.
\newblock {\em arXiv preprint arXiv:2104.12533}, 2021.

\bibitem{choromanski2020rethinking}
Krzysztof Choromanski, Valerii Likhosherstov, David Dohan, Xingyou Song,
  Andreea Gane, Tamas Sarlos, Peter Hawkins, Jared Davis, Afroz Mohiuddin,
  Lukasz Kaiser, et~al.
\newblock Rethinking attention with performers.
\newblock {\em arXiv preprint arXiv:2009.14794}, 2020.

\bibitem{twins}
Xiangxiang Chu, Zhi Tian, Yuqing Wang, Bo Zhang, Haibing Ren, Xiaolin Wei,
  Huaxia Xia, and Chunhua Shen.
\newblock Twins: Revisiting the design of spatial attention in vision
  transformers.
\newblock {\em arXiv preprint arXiv:2104.13840}, 1(2):3, 2021.

\bibitem{chu2021fairnas}
Xiangxiang Chu, Bo Zhang, and Ruijun Xu.
\newblock Fairnas: Rethinking evaluation fairness of weight sharing neural
  architecture search.
\newblock In {\em Proceedings of the IEEE/CVF International Conference on
  Computer Vision}, pages 12239--12248, 2021.

\bibitem{deng2009imagenet}
Jia Deng, Wei Dong, Richard Socher, Li-Jia Li, Kai Li, and Li Fei-Fei.
\newblock Imagenet: A large-scale hierarchical image database.
\newblock In {\em 2009 IEEE conference on computer vision and pattern
  recognition}, pages 248--255. Ieee, 2009.

\bibitem{dosovitskiy2020image}
Alexey Dosovitskiy, Lucas Beyer, Alexander Kolesnikov, Dirk Weissenborn,
  Xiaohua Zhai, Thomas Unterthiner, Mostafa Dehghani, Matthias Minderer, Georg
  Heigold, Sylvain Gelly, et~al.
\newblock An image is worth 16x16 words: Transformers for image recognition at
  scale.
\newblock {\em arXiv preprint arXiv:2010.11929}, 2020.

\bibitem{elbayad2019depth}
Maha Elbayad, Jiatao Gu, Edouard Grave, and Michael Auli.
\newblock Depth-adaptive transformer.
\newblock {\em arXiv preprint arXiv:1910.10073}, 2019.

\bibitem{mvit}
Haoqi Fan, Bo Xiong, Karttikeya Mangalam, Yanghao Li, Zhicheng Yan, Jitendra
  Malik, and Christoph Feichtenhofer.
\newblock Multiscale vision transformers.
\newblock {\em arXiv preprint arXiv:2104.11227}, 2021.

\bibitem{frankle2018lottery}
Jonathan Frankle and Michael Carbin.
\newblock The lottery ticket hypothesis: Finding sparse, trainable neural
  networks.
\newblock {\em arXiv preprint arXiv:1803.03635}, 2018.

\bibitem{guo2020single}
Zichao Guo, Xiangyu Zhang, Haoyuan Mu, Wen Heng, Zechun Liu, Yichen Wei, and
  Jian Sun.
\newblock Single path one-shot neural architecture search with uniform
  sampling.
\newblock In {\em European Conference on Computer Vision}, pages 544--560.
  Springer, 2020.

\bibitem{tnt}
Kai Han, An Xiao, Enhua Wu, Jianyuan Guo, Chunjing Xu, and Yunhe Wang.
\newblock Transformer in transformer.
\newblock {\em arXiv preprint arXiv:2103.00112}, 2021.

\bibitem{han2015deep}
Song Han, Huizi Mao, and William~J Dally.
\newblock Deep compression: Compressing deep neural networks with pruning,
  trained quantization and huffman coding.
\newblock {\em arXiv preprint arXiv:1510.00149}, 2015.

\bibitem{he2021masked}
Kaiming He, Xinlei Chen, Saining Xie, Yanghao Li, Piotr Doll{\'a}r, and Ross
  Girshick.
\newblock Masked autoencoders are scalable vision learners.
\newblock {\em arXiv preprint arXiv:2111.06377}, 2021.

\bibitem{hinton2015distilling}
Geoffrey Hinton, Oriol Vinyals, and Jeff Dean.
\newblock Distilling the knowledge in a neural network.
\newblock {\em arXiv preprint arXiv:1503.02531}, 2015.

\bibitem{hu2018relation}
Han Hu, Jiayuan Gu, Zheng Zhang, Jifeng Dai, and Yichen Wei.
\newblock Relation networks for object detection.
\newblock In {\em Proceedings of the IEEE conference on computer vision and
  pattern recognition}, pages 3588--3597, 2018.

\bibitem{lvvit}
Zihang Jiang, Qibin Hou, Li Yuan, Daquan Zhou, Xiaojie Jin, Anran Wang, and
  Jiashi Feng.
\newblock Token labeling: Training a 85.5\% top-1 accuracy vision transformer
  with 56m parameters on imagenet.
\newblock {\em arXiv preprint arXiv:2104.10858}, 2021.

\bibitem{kay2017kinetics}
Will Kay, Joao Carreira, Karen Simonyan, Brian Zhang, Chloe Hillier, Sudheendra
  Vijayanarasimhan, Fabio Viola, Tim Green, Trevor Back, Paul Natsev, et~al.
\newblock The kinetics human action video dataset.
\newblock {\em arXiv preprint arXiv:1705.06950}, 2017.

\bibitem{li2021mst}
Zhaowen Li, Zhiyang Chen, Fan Yang, Wei Li, Yousong Zhu, Chaoyang Zhao, Rui
  Deng, Liwei Wu, Rui Zhao, Ming Tang, et~al.
\newblock Mst: Masked self-supervised transformer for visual representation.
\newblock {\em arXiv preprint arXiv:2106.05656}, 2021.

\bibitem{liu2018darts}
Hanxiao Liu, Karen Simonyan, and Yiming Yang.
\newblock Darts: Differentiable architecture search.
\newblock {\em arXiv preprint arXiv:1806.09055}, 2018.

\bibitem{liu2021swin}
Ze Liu, Yutong Lin, Yue Cao, Han Hu, Yixuan Wei, Zheng Zhang, Stephen Lin, and
  Baining Guo.
\newblock Swin transformer: Hierarchical vision transformer using shifted
  windows.
\newblock {\em arXiv preprint arXiv:2103.14030}, 2021.

\bibitem{liu2021post}
Zhenhua Liu, Yunhe Wang, Kai Han, Siwei Ma, and Wen Gao.
\newblock Post-training quantization for vision transformer.
\newblock {\em arXiv preprint arXiv:2106.14156}, 2021.

\bibitem{mahmood2021robustness}
Kaleel Mahmood, Rigel Mahmood, and Marten Van~Dijk.
\newblock On the robustness of vision transformers to adversarial examples.
\newblock {\em arXiv preprint arXiv:2104.02610}, 2021.

\bibitem{naseer2021intriguing}
Muzammal Naseer, Kanchana Ranasinghe, Salman Khan, Munawar Hayat, Fahad~Shahbaz
  Khan, and Ming-Hsuan Yang.
\newblock Intriguing properties of vision transformers.
\newblock {\em arXiv preprint arXiv:2105.10497}, 2021.

\bibitem{pan2021ia}
Bowen Pan, Rameswar Panda, Yifan Jiang, Zhangyang Wang, Rogerio Feris, and Aude
  Oliva.
\newblock Ia-red $^ 2$: Interpretability-aware redundancy reduction for vision
  transformers.
\newblock {\em arXiv preprint arXiv:2106.12620}, 2021.

\bibitem{pointformer}
Xuran Pan, Zhuofan Xia, Shiji Song, Li~Erran Li, and Gao Huang.
\newblock 3d object detection with pointformer.
\newblock In {\em Proceedings of the IEEE/CVF Conference on Computer Vision and
  Pattern Recognition}, pages 7463--7472, 2021.

\bibitem{raghu2021vision}
Maithra Raghu, Thomas Unterthiner, Simon Kornblith, Chiyuan Zhang, and Alexey
  Dosovitskiy.
\newblock Do vision transformers see like convolutional neural networks?
\newblock {\em arXiv preprint arXiv:2108.08810}, 4, 2021.

\bibitem{rao2021dynamicvit}
Yongming Rao, Wenliang Zhao, Benlin Liu, Jiwen Lu, Jie Zhou, and Cho-Jui Hsieh.
\newblock Dynamicvit: Efficient vision transformers with dynamic token
  sparsification.
\newblock {\em arXiv preprint arXiv:2106.02034}, 2021.

\bibitem{ryoo2021tokenlearner}
Michael~S Ryoo, AJ Piergiovanni, Anurag Arnab, Mostafa Dehghani, and Anelia
  Angelova.
\newblock Tokenlearner: What can 8 learned tokens do for images and videos?
\newblock {\em arXiv preprint arXiv:2106.11297}, 2021.

\bibitem{cait}
Hugo Touvron Matthieu Cord~Alexandre Sablayrolles and Gabriel
  Synnaeve~Herv{\'e} J{\'e}gou.
\newblock Going deeper with image transformers.

\bibitem{song2021dynamic}
Lin Song, Songyang Zhang, Songtao Liu, Zeming Li, Xuming He, Hongbin Sun, Jian
  Sun, and Nanning Zheng.
\newblock Dynamic grained encoder for vision transformers.
\newblock In {\em Thirty-Fifth Conference on Neural Information Processing
  Systems}, 2021.

\bibitem{tan2019efficientnet}
Mingxing Tan and Quoc Le.
\newblock Efficientnet: Rethinking model scaling for convolutional neural
  networks.
\newblock In {\em International Conference on Machine Learning}, pages
  6105--6114. PMLR, 2019.

\bibitem{tang2021patch}
Yehui Tang, Kai Han, Yunhe Wang, Chang Xu, Jianyuan Guo, Chao Xu, and Dacheng
  Tao.
\newblock Patch slimming for efficient vision transformers.
\newblock {\em arXiv preprint arXiv:2106.02852}, 2021.

\bibitem{deit}
Hugo Touvron, Matthieu Cord, Matthijs Douze, Francisco Massa, Alexandre
  Sablayrolles, and Herv{\'e} J{\'e}gou.
\newblock Training data-efficient image transformers \& distillation through
  attention.
\newblock In {\em International Conference on Machine Learning}, pages
  10347--10357. PMLR, 2021.

\bibitem{vaswani2017attention}
Ashish Vaswani, Noam Shazeer, Niki Parmar, Jakob Uszkoreit, Llion Jones,
  Aidan~N Gomez, {\L}ukasz Kaiser, and Illia Polosukhin.
\newblock Attention is all you need.
\newblock In {\em Advances in neural information processing systems}, pages
  5998--6008, 2017.

\bibitem{wang2018non}
Xiaolong Wang, Ross Girshick, Abhinav Gupta, and Kaiming He.
\newblock Non-local neural networks.
\newblock In {\em Proceedings of the IEEE conference on computer vision and
  pattern recognition}, pages 7794--7803, 2018.

\bibitem{wang2021not}
Yulin Wang, Rui Huang, Shiji Song, Zeyi Huang, and Gao Huang.
\newblock Not all images are worth 16x16 words: Dynamic vision transformers
  with adaptive sequence length.
\newblock {\em arXiv preprint arXiv:2105.15075}, 2021.

\bibitem{wang2020glance}
Yulin Wang, Kangchen Lv, Rui Huang, Shiji Song, Le Yang, and Gao Huang.
\newblock Glance and focus: a dynamic approach to reducing spatial redundancy
  in image classification.
\newblock {\em arXiv preprint arXiv:2010.05300}, 2020.

\bibitem{wu2021cvt}
Haiping Wu, Bin Xiao, Noel Codella, Mengchen Liu, Xiyang Dai, Lu Yuan, and Lei
  Zhang.
\newblock Cvt: Introducing convolutions to vision transformers.
\newblock {\em arXiv preprint arXiv:2103.15808}, 2021.

\bibitem{xiao2021early}
Tete Xiao, Mannat Singh, Eric Mintun, Trevor Darrell, Piotr Doll{\'a}r, and
  Ross Girshick.
\newblock Early convolutions help transformers see better.
\newblock {\em arXiv preprint arXiv:2106.14881}, 2021.

\bibitem{coat}
Weijian Xu, Yifan Xu, Tyler Chang, and Zhuowen Tu.
\newblock Co-scale conv-attentional image transformers.
\newblock {\em arXiv preprint arXiv:2104.06399}, 2021.

\bibitem{yang2020transpose}
Sen Yang, Zhibin Quan, Mu Nie, and Wankou Yang.
\newblock Transpose: Towards explainable human pose estimation by transformer.
\newblock {\em arXiv preprint arXiv:2012.14214}, 2020.

\bibitem{yin2020disentangled}
Minghao Yin, Zhuliang Yao, Yue Cao, Xiu Li, Zheng Zhang, Stephen Lin, and Han
  Hu.
\newblock Disentangled non-local neural networks.
\newblock In {\em European Conference on Computer Vision}, pages 191--207.
  Springer, 2020.

\bibitem{yu2019universally}
Jiahui Yu and Thomas~S Huang.
\newblock Universally slimmable networks and improved training techniques.
\newblock In {\em Proceedings of the IEEE/CVF International Conference on
  Computer Vision}, pages 1803--1811, 2019.

\bibitem{yu2020bignas}
Jiahui Yu, Pengchong Jin, Hanxiao Liu, Gabriel Bender, Pieter-Jan Kindermans,
  Mingxing Tan, Thomas Huang, Xiaodan Song, Ruoming Pang, and Quoc Le.
\newblock Bignas: Scaling up neural architecture search with big single-stage
  models.
\newblock In {\em European Conference on Computer Vision}, pages 702--717.
  Springer, 2020.

\bibitem{yu2018slimmable}
Jiahui Yu, Linjie Yang, Ning Xu, Jianchao Yang, and Thomas Huang.
\newblock Slimmable neural networks.
\newblock {\em arXiv preprint arXiv:1812.08928}, 2018.

\bibitem{poseformer}
Ce Zheng, Sijie Zhu, Matias Mendieta, Taojiannan Yang, Chen Chen, and Zhengming
  Ding.
\newblock 3d human pose estimation with spatial and temporal transformers.
\newblock {\em arXiv preprint arXiv:2103.10455}, 2021.

\bibitem{zhu2020deformable}
Xizhou Zhu, Weijie Su, Lewei Lu, Bin Li, Xiaogang Wang, and Jifeng Dai.
\newblock Deformable detr: Deformable transformers for end-to-end object
  detection.
\newblock {\em arXiv preprint arXiv:2010.04159}, 2020.

\bibitem{fbnnas}
Yichen Zhu, Du Jie, Zhu Yuqin, Wang Yi, Ou Zhicai, Feng Feifei, and Tang Jian.
\newblock Training batchnorm only in neural architecture search and beyond.
\newblock {\em arXiv preprint arXiv:2112.00265}, 2021.

\bibitem{zhu2021student}
Yichen Zhu and Yi Wang.
\newblock Student customized knowledge distillation: Bridging the gap between
  student and teacher.
\newblock In {\em Proceedings of the IEEE/CVF International Conference on
  Computer Vision}, pages 5057--5066, 2021.

\bibitem{zhu2019resizable}
Yichen Zhu, Xiangyu Zhang, Tong Yang, and Jian Sun.
\newblock Resizable neural networks.
\newblock 2019.

\bibitem{zoph2016neural}
Barret Zoph and Quoc~V Le.
\newblock Neural architecture search with reinforcement learning.
\newblock {\em arXiv preprint arXiv:1611.01578}, 2016.

\end{thebibliography}
}

\end{document}